\documentclass[10pt,twocolumn,letterpaper]{article}

\usepackage[pagenumbers]{cvpr} 

\usepackage{graphicx}
\usepackage{amsmath}
\usepackage{amssymb}
\usepackage{booktabs}
\usepackage[flushleft]{threeparttable}
\usepackage[capitalize]{cleveref}
\crefname{section}{Sec.}{Secs.}
\Crefname{section}{Section}{Sections}
\Crefname{table}{Table}{Tables}
\crefname{table}{Tab.}{Tabs.}

\def\cvprPaperID{4035} 
\def\confName{CVPR}
\def\confYear{2022}

\usepackage{overpic}
\usepackage{enumitem} 
\usepackage{overpic} 
\usepackage{color}

\definecolor{turquoise}{cmyk}{0.65,0,0.1,0.3}
\definecolor{purple}{rgb}{0.65,0,0.65}
\definecolor{dark_green}{rgb}{0, 0.5, 0}
\definecolor{orange}{rgb}{0.8, 0.6, 0.2}
\definecolor{red}{rgb}{0.8, 0.2, 0.2}
\definecolor{darkred}{rgb}{0.6, 0.1, 0.05}
\definecolor{blueish}{rgb}{0.0, 0.3, .6}
\definecolor{light_gray}{rgb}{0.7, 0.7, .7}
\definecolor{pink}{rgb}{1, 0, 1}
\definecolor{greyblue}{rgb}{0.25, 0.25, 1}

\usepackage{blindtext}

\renewcommand{\paragraph}[1]{\vspace{1em}\noindent\textbf{#1}.}
\begin{document}
\title{Multi-modal Domain Adaptation for REG via Relation Transfer}
\author{Yifan Ding \hspace{2cm} Liqiang Wang \\
Department of Computer Science, University of Central Florida\\
{\tt\small yf.ding@knights.ucf.edu, lwang@cs.ucf.edu}
\and
Boqing Gong \\
Google\\
{\tt\small  bgong@google.com}
}

\maketitle
\begin{abstract}
Domain adaptation, which aims to transfer knowledge between domains, has been well studied in many areas such as image classification and object detection. However, for multi-modal tasks, conventional approaches rely on large-scale pre-training. But due to the difficulty of acquiring multi-modal data, large-scale pre-training is often impractical. Therefore, domain adaptation, which can efficiently utilize the knowledge from different datasets (domains), is crucial for multi-modal tasks. In this paper, we focus on the Referring Expression Grounding (REG) task, which is to localize an image region described by a natural language expression. Specifically, we propose a novel approach to effectively transfer multi-modal knowledge through a specially relation-tailored approach for the REG problem. Our approach tackles the multi-modal domain adaptation problem by simultaneously enriching inter-domain relations and transferring relations  between domains. Experiments show that our proposed approach significantly improves the transferability of multi-modal domains and enhances adaptation performance in the REG problem.






\end{abstract}
\section{Introduction}


Domain adaptation aims to mitigate the discrepancy between domains such that a model learned from the source domain can be well generalized to the target domain. Recent methods in domain adaptation mainly study single-modality tasks, such as image classification~\cite{you2019universal,long2017conditional}, semantic segmentation~\cite{zhang2017curriculum,du2019ssf}, object detection~\cite{inoue2018cross,hsu2020progressive}, and natural language processing~\cite{ramponi2020neural,laparra2020rethinking}. However, when it comes to multi-modal data, the current approaches mainly rely on  pre-training  on large-scale multi-modal datasets 
followed by fine-tuning on downstream tasks~\cite{chen2020uniter,lu2019vilbert,su2019vl}. Such plain transfer learning methods are time-consuming and resource-consuming.

\begin{figure}[t]
\begin{center}
   \includegraphics[width=0.95\linewidth]{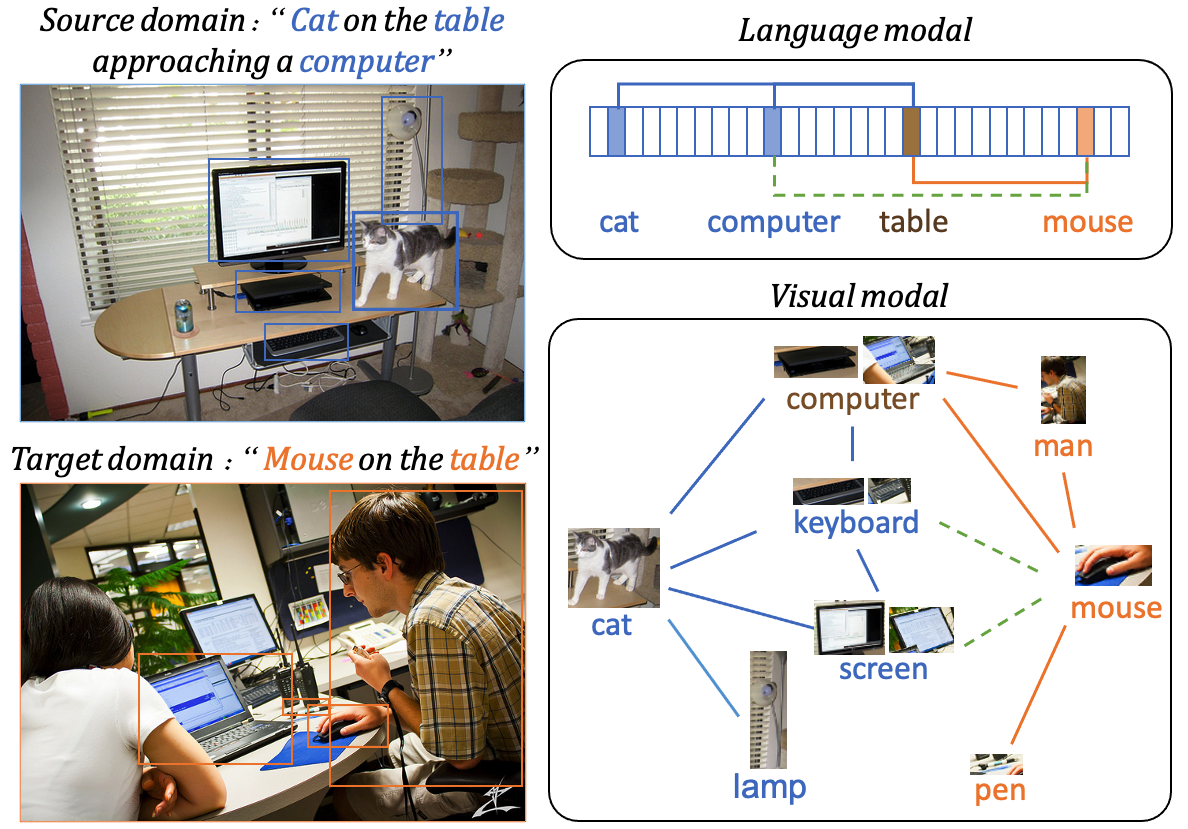}
\end{center}
   \caption{An example of relation transfer. Relations learned from the source domain are shown in blue, and relations learned from the target domain are shown in orange in both the language and visual modals. Relations shared between two domains are denoted with brown color. New relations created after transferring from source domain to target domain are shown in green dash lines.  }
\label{fig:example}
\end{figure}

In this paper, we investigate multi-modal knowledge transfer and propose a new effective approach in a domain adaptation manner. Specifically, we focus on the referring expression grounding (REG)~\cite{kazemzadeh2014referitgame} problem. REG aims to localize an image region described by a natural language expression. As a multi-modal problem, REG requires high-level language comprehension and object localization, which make collecting and labeling REG data be expensive and time-consuming. To make up for the limited training data, \textit{pretrain-then-transfer} learning approaches are widely used in recent research studies~\cite{chen2020uniter,lu2019vilbert,su2019vl}. It mainly relies on a ``vision-and-language" feature extractor, which is often pre-trained with large-scale and usually roughly annotated  ``vision-and-language" datasets (\eg the Conceptual Captions dataset~\cite{sharma2018conceptual}) and then fine-tuned together with the downstream classifier. However, there are several limitations in such an approach. First, there is a gap between the general ``vision-and-language" feature extractor and the REG task, especially when the most important comprehension part is missing in pre-training due to the annotation limits. Secondly, the feature extractor (\eg transformer-based) is usually heavyweight, which results in large model size, big pre-training dataset, and long pre-training time. It also causes a burdened and slow fine-tuning process. 

The goal of this paper is to effectively transfer knowledge between REG datasets to avoid the above problems. However, it is not an easy task. First, to capture  the variance among different types of referring expressions, a widely-used approach is  modular networks, which contain different visual modules processing different information such as name, attributes, location, or relations to other objects in the referring expression~\cite{yu2018mattnet}. Such modular networks lack a joint vision-language embedding~\cite{yu2018mattnet,liu2019improving,gupta2020contrastive}. Thus regular domain adaptation methods that focus on aligning features in two domains ~\cite{ganin2015unsupervised,long2017conditional,cai2019learning} cannot be applied . Moreover, enforcing domain-invariance for each modality (language or vision) individually fails to capture their interplay. Besides, the domain gap between two REG domains stems from multiple factors, such as visual styles, biases in the bounding box annotations, and textual mismatches in styles and dictionaries, making it difficult to straightforwardly extend existing single modality adaptation methods to the REG task. 


To address the above problems, we propose to extract and transfer language/objects features in a relation-based manner rather than aligning the features. Figure~\ref{fig:example} shows an example. In this example, relations between words/objects in the source domain are learned in the pre-training. These learned relations are universal and transferable (\eg computer always sit on table in both domains). Besides, they also help uncovering underlying relations in the target task. For instance, in the language modal, from source domain we learned a relation (``computer'', ``table''); in the target domain, ``mouse'' is only connected with ``table" but an underlying connection (``mouse'', ``computer'') could be discovered. In the visual modal, we have similar cases. Although our relation-based domain adaptation approach is specially tailored for the REG problem, it could be extended to other multi-modal tasks.  To the best of our knowledge, we are the first to solve the problem from this angle. Our approach aims to discover and maintain the relations between language/objects and mitigate the multi-modal discrepancy between the source and target domains. 

From the language aspect, the discrepancy pivots on the difference between source and target vocabularies. Since the source and target usually share parts of their vocabularies and each has its own domain-specific words, our task is to transfer learned embeddings from source-private and shared vocabularies to target-private vocabulary. To achieve it, we design the cross-attention embedding strategy, through which the two domains share common knowledge in the shared vocabulary, meanwhile maintaining sufficient expressive power to learn  domain-specific knowledge.




Among the visual side, in the REG model, the relation module that handles surrounding objects of a candidate object plays an important role in comprehension. Through the relation module, the REG model recognizes objects in the same category but have different surroundings (\eg ``man in the middle of three men'').
Previous approaches use geographical distance to define the surrounding objects and pick up one with the highest matching score to represent the environment of the candidate objects~\cite{yu2018mattnet,liu2019improving}. In our approach, however, we propose to re-build the relation modular with Graph Neural Networks (GNN)~\cite{scarselli2008graph}. GNN could model the relations between objects in terms of graphs, which better describe the surrounding environment of a candidate object. Besides, GNNs could utilize the co-occurrences between objects in the source dataset. That also improves the features while transferring to the target dataset. In addition, we extend scene graph~\cite{xu2017scene} to  enrich the graphs with semantic relations between objects.


Our main contributions are as follows. First, we formulate domain adaption for the REG problem as a multi-modal transfer learning task, which enriches the transfer learning techniques in the vision-language research. 
Secondly, we design a new relation-based adaptation method for the REG problem, which enables better knowledge transfer to new REG dataset. Our method results in a fast and lightweight transfer compared with traditional expensive \textit{pretrain-then-transfer} learning schema. 
Thirdly, the proposed multi-modal relation-based domain adaption method significantly outperforms other approaches for the REG problem. 

\section{Related Work}
\label{sec:related}

Our work is at the intersection of several subareas in the framework of deep learning. In this section, we discuss related work on REG, domain adaptation, and multi-modal adaptation for REG.
\begin{figure*}[t]
\begin{center}
   \includegraphics[width=0.85\linewidth]{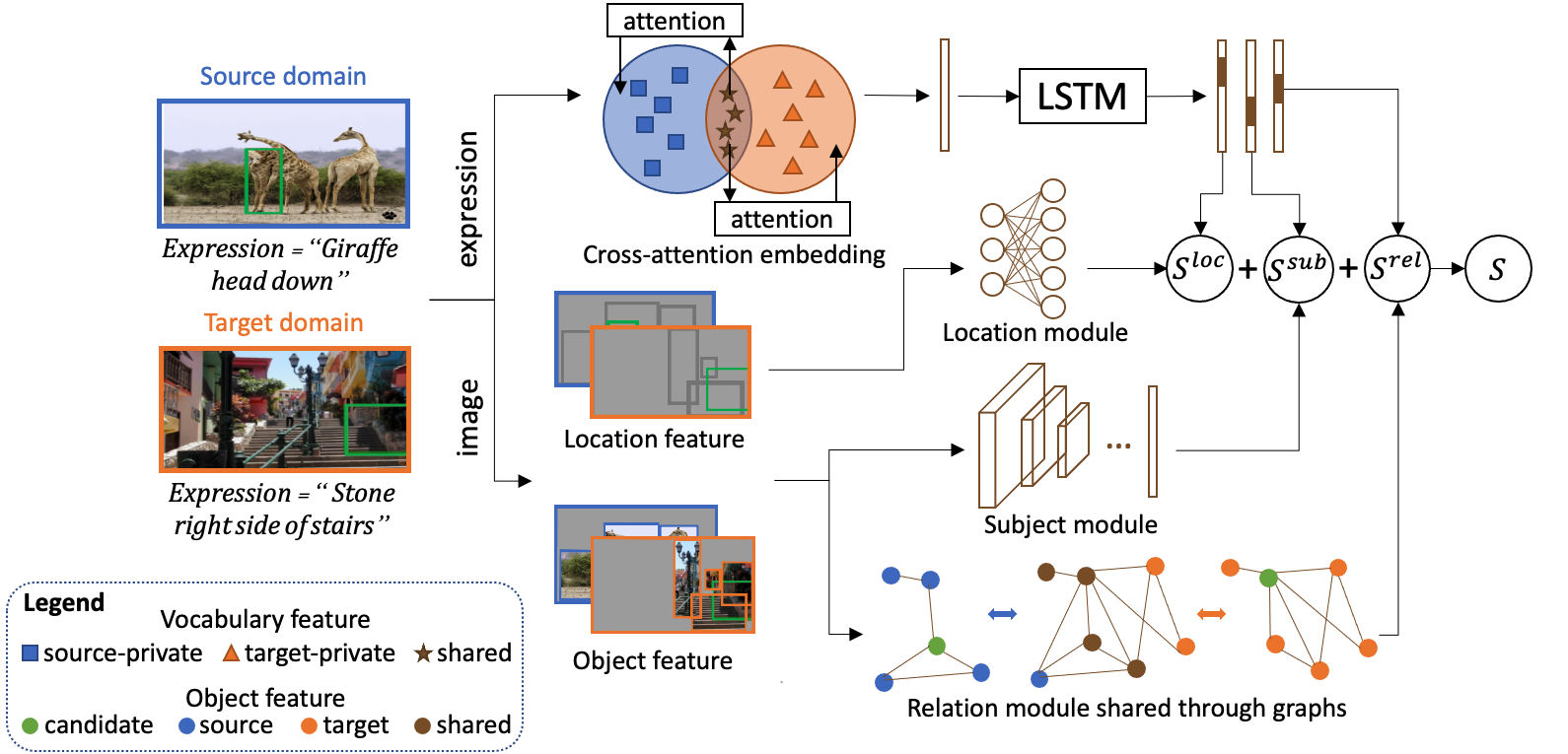}
\end{center}
   \caption{The architecture of our proposed approach. 
   The model is composed of a language module in which the cross-attention embedding is applied and three visual modules (\ie, location, subject, and relation modules). The final matching score $S$ between expression and visual modules is the weighted sum of three modules $S_{loc}$, $S_{sub}$ and $S_{rel}$.  } 
\label{fig:overview}
\end{figure*}

\textbf{Referring expression grounding.} Like most vision-language tasks~\cite{antol2015vqa,wang2016learning}, REG also relies on multi-modal embeddings to bridge the semantic gap between visual and textual contents.
However, when data distributions embody complex multi-modal structures, general domain adaptation methods may fail to capture such multi-modal structures for a discriminative alignment of distributions. Therefore, in most cases, either training from scratch~\cite{yu2018mattnet,liu2019improving} or large-scale pre-trained models are used to transfer knowledge from large-scale vision-language dataset to specific vision-language tasks~\cite{chen2020uniter,lu2019vilbert,su2019vl}. In this paper, we adopt a modular structure proposed in MAttNet~\cite{yu2018mattnet}. It has a language module and three visual modules that process the visual feature, location feature and the surrounding object feature of the candidate object, respectively. We reformulate the language module to process the relations between words and the visual relation module to model the visual object relations. 


\textbf{Domain adaptation.} Domain adaptation aims to mitigate the discrepancy of distributions between source domain and target domain data. 
The main methods in domain adaptation are to minimize the statistical distances between source and target feature distributions. 
These methods usually take advantage of adversarial training 
~\cite{mirza2014conditional} and include a domain classifier that discriminates between the source and the target domains during training~\cite{ganin2015unsupervised,tzeng2017adversarial}. In that direction, some recent methods condition the domain discriminator to facilitate more accurate feature adaptation within each category~\cite{long2017conditional,hu2020unsupervised}. However, the adversarial training procedure tends to be unstable and may cause performance drop. Strengthening Lipschitz continuity in the target distribution is a method that  avoids adversarial training yet still supports domain adaptation~\cite{shu2018dirt,mao2019virtual,cai2019learning}. 

\textbf{Domain adaptation for referring expression grounding.} Since general domain adaptation methods cannot be applied to the multi-modal problem directly, several algorithms have been proposed to align distributions of features and classes via separate domain discriminators~\cite{hoffman2016fcns,chen2017no,tsai2018learning}. In particular for the REG task, Liu et al. propose to transfer concept from
auxiliary classification data (images from new categories) and context inheritance from REG data to ground new objects~\cite{liu2020transferrable}. In our approach, Graph Convolutional Network (GCN)~\cite{kipf2016semi} is used to transfer knowledge from source to target domains. Our approach is different from the transfer learning of graph network~\cite{lee2017transfer,zhu2020transfer}, as our main task is to transfer the relations, not the graph itself. Besides, pivot-based methods~\cite{blitzer2006domain,ziser2016neural,ziser2018pivot}, which recognize the frequent features in the source and target domains, also inspire our cross-attention dictionary method.

\section{Multi-modal Domain Adaptation for REG}
\label{sec:approach}
Our main approach is described in this section. The framework is within supervised domain adaptation, where the model has access to a small amount of labeled data from the source and target domains. This allows us to pre-train the model in the source domain and then fine-tune it in the target domain. 

Different from previous REG models~\cite{yu2018mattnet,liu2019improving,gupta2020contrastive}, our model transfers language and visual features based on the word relations and visual object relations. Our proposed model results in less discrepancy and preserves more target domain oriented  information during the transfer. 

\subsection{Cross-attention Word Embedding}
\label{subsec:WordEmbedding}

In the language module, the first step is to encode words in the expression sentence to fixed-length feature vectors. Under the domain adaptation scenario, the source and target domains are likely to have different vocabularies. For example, the RefCOCO dataset~\cite{yu2016modeling} has 1999 words while the RefCOCOg dataset~\cite{mao2016generation} has 3349 words. We denote the vocabulary that appears in both the source and target dataset as ``shared vocabulary'', and the vocabulary that only appears in the source or target datasets as ``source-private'' and ``target-private" vocabulary, respectively. 
Our goal is to transfer as much information from the source domain to the target domain as possible, for which the shared words are pivotal, it is not sufficient. We conjecture that not only the shared words can be transferred between domains, the correlation among words also transfers. We propose an attention mechanism to model the correlations between words, in which the embeddings of domain private vocabularies are learned from both the training data and the attended shared vocabulary embeddings. By the attention machinism, we build the relationship between shared vocabulary and private vocabularies, which also connects the source and target private words indirectly.   
Two benefits are brought by the relation between shared and private vocabularies :  1) words in the shared vocabulary are usually more widely used and point to more general objects and represent more common semantics which enrich the representation of private words with more general and recognizable features , and 2) the embeddings of shared vocabulary turn to be more domain-invariant to represent privates from two domains.


In the cross-attention embedding, we denote the embedding matrix by $D \in \mathbb{R}^{d \times n }$ for a dictionary of $n$ words, each of which is embedded to a $d$-dimensional vector. We denote the source and target vocabulary embeddings by $D_{S} $ and $ D_{T}$, the shared and two private vocabulary embeddings by $ D_{S \cap T}, D_{S\backslash T}$ and $D_{T\backslash S}$, respectively.  Thus, $D = D_{S \cap T} \cup D_{S\backslash T} \cup D_{T\backslash S}$. We also have $D_{S} = D_{S \cap T} \cup D_{S\backslash T} $ and $D_{T} = D_{S \cap T} \cup D_{T\backslash S}$. $D_{S \cap T}$ is learnt from the dataset, while $D_{S\backslash T}$ and $D_{T\backslash S}$ are attended by $D_{S \cap T}$ with learnable matrices $W_{S \cap T \rightarrow S \backslash T}$ and $W_{S \cap T \rightarrow T \backslash S}$, which model the relations between the shared vocabulary and the source and target private vocabulary, respectively.  These embedding matrices $D$ and attention matrices $W$ are trained together with the language attention module in the REG model, as shown in Figure \ref{fig:overview}.

\begin{figure}[t]
\begin{center}
   \includegraphics[width=0.85\linewidth]{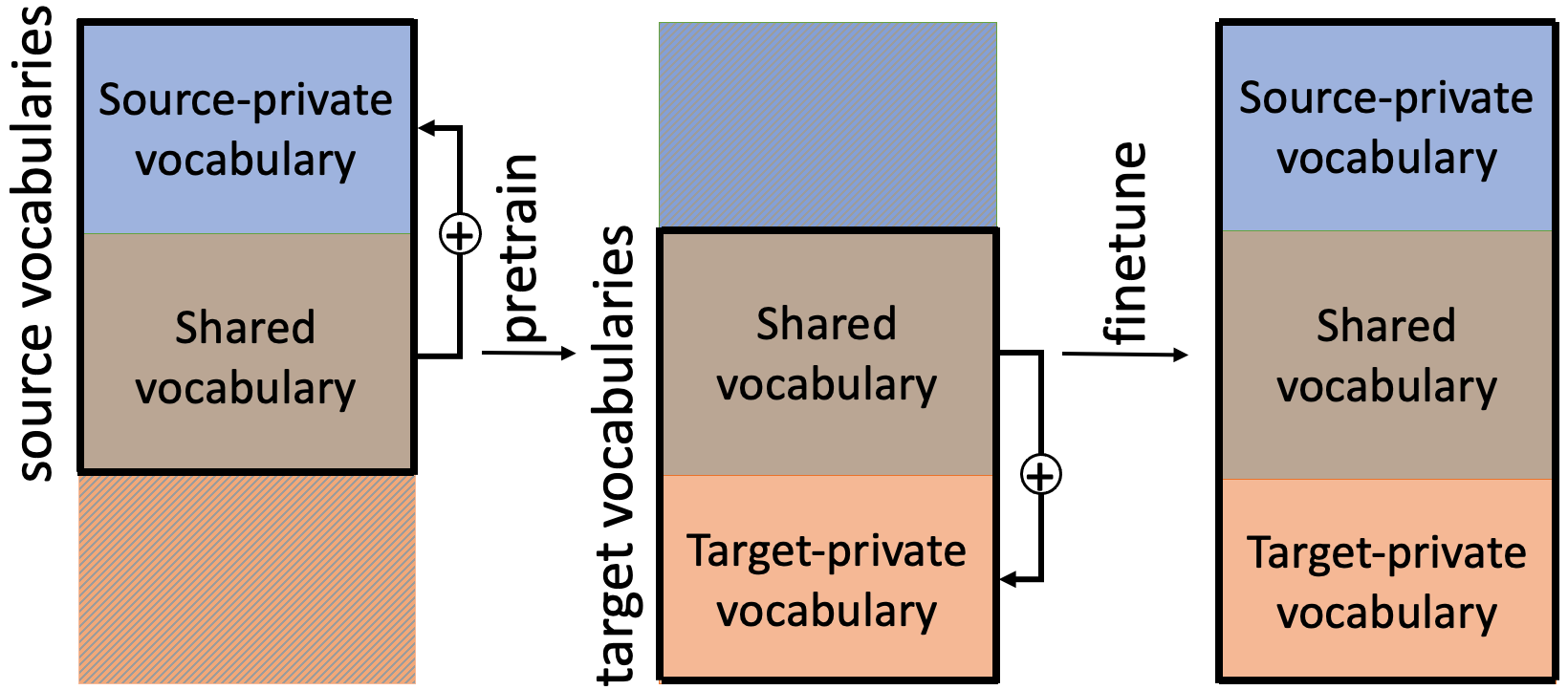}
\end{center}
   \caption{Illustration of the workflow of cross-attention word embedding. During the pre-training, the attention (denoted by $\oplus$) is learnt from shared vocabulary $D_{S \cap T}$ to source-private vocabulary $D_{S\backslash T}$. In the fine-tuning, the attention is learnt from shared vocabulary $D_{S \cap T}$ to target-private vocabulary $D_{T\backslash S}$.}
\label{fig:att}
\end{figure}

During pretraining, the update during forward pass is:
\begin{align}\begin{aligned}
    \overline{D_{S \backslash T}} &= D_{S \backslash T} + D_{S \cap T} \cdot \textrm{softmax}( W_{S\cap T \rightarrow S\backslash T}  ),  \\
    \overline{D_{S \cap T}} &= D_{S \cap T}, 
\end{aligned}\end{align}
where $\cdot$ is matrix multiplication. $W_{S \cap T \rightarrow S \backslash T} $ is the learnable weights of shape $n_{S \cap T} \times n_ {S \backslash T }$. In every forward pass, the private vocabulary embedding $D_{S\backslash T}$ is updated (denoted as $\overline{D_{S \backslash T}}$) with the sum of original embedding $D_{S\backslash T}$ and $D_{S\cap T}$ weighted by a softmax version of $W_{S \cap T \rightarrow S \backslash T} $.  
After back-propagation, we update $D$ which are trainable weights with learning rate $\eta$ :
\begin{align}
    D_{S\backslash T} \leftarrow D_{S\backslash T} - \eta \frac{\partial Loss}{\partial D_{S\backslash T}}, ~~
\nonumber D_{S\cap T} \leftarrow D_{S\cap T} - \eta \frac{\partial Loss}{\partial D_{S\cap T}} \nonumber
\end{align}

During finetuning, $D_{T\backslash S}$ is updated with the target data and the learned embeddings $D_{S\cap T}$ from the source domain:
\begin{align}\begin{aligned}
    \overline{D_{T \backslash S}} &= D_{T \backslash S} + 
    D_{S \cap T} \cdot \textrm{softmax}({ W_{S\cap T \rightarrow T\backslash S}}  ), \\
    \overline{D_{S \cap T}} &= D_{S \cap T},
\end{aligned}\end{align}
$W_{S \cap T \rightarrow T \backslash S} $ is of shape $n_{S \cap T} \times n_ {T \backslash S }$. $D_{S \backslash T}$ and $D_{T \backslash S}$ are also updated in propagation. 

Since private words are usually less common and have less training samples, this mechanism improves the embedding quality in both $D_{S\backslash T}$ and $D_{S\backslash T}$.. At the same time, the embeddings in shared vocabulary also become more domain invariant by learning to represent private words from the source and target domains.

With the  embedding matrix $D$, we get domain invariant word embedding in pre-training and fine-tuning. After that, we apply a bi-directional LSTM~~\cite{graves2005bidirectional} to encode the whole expression $R= \{u_t\}^T_{t=1}$:
\begin{align}\begin{aligned}
e_t &= D(u_t), \\
h_t &= \textrm{Bi-LSTM}(e_t, h_{t-1}),\\
\end{aligned}\end{align}
$H= \{h_t\}^T_{t=1}$ is the expression feature.






\subsection{Object Relation Transfer via GNNs }
In visual module, we also conduct domain adaptation by discovering transferable relationships. For REG models, pairwise data are used in the training, where each pair of data is consisted of a sentence of expression (processed in the language module) and a candidate object. Usually a Convolutional Neural Network~\cite{krizhevsky2012imagenet} is implemented to process the visual feature (subject module) while a multi-layer perceptron without convolutions is used to process the location feature (location module) of the object. However, these two modules are not enough to location a target object. For example, in the expression ``first giraffe on left'', we have to locate the surrounding objects of each giraffe and only the one with no other giraffe on its left is the correct one. Here, the co-occurrence of objects also helps locate the target object, even when the expression didn't mention them. 

Based on the above observations, we propose to describe the relations between objects in an image with graphs. 
Furthermore, we merge graphs from source and target domains into one during learning via Graph Neural Networks (GNN)~\cite{scarselli2008graph}. GNNs typically aggregate information from neighbor nodes to embedded information into a hidden space regardless of which domain the node is from. It utilizes the co-occurrence information between objects. Therefore when transferring to a new dataset, the model still benefits from the learned connections between objects in the source dataset. At the same time, the GNNs are also updated with the new data coming from the target dataset. 

\begin{figure}[t]
\begin{center}
  \includegraphics[width=0.75\linewidth]{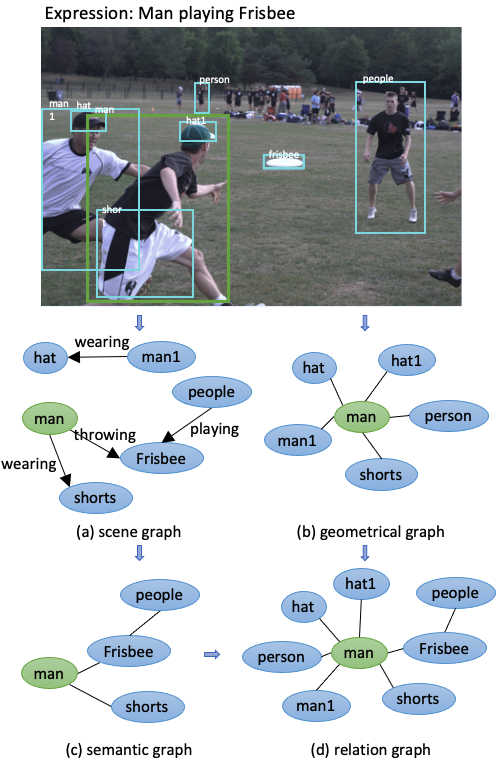}
\end{center}
  \caption{An example of relation graph generalization.}
\label{fig:graph}
\end{figure}
\subsubsection{Relation Graph Generalization} 

Our first step is to create the relation graph of candidate object $v$ by $\mathcal{G}_v = (\mathcal{V}_v,\mathcal{A}_v)$, where $\mathcal{V}_v$ is the set of nodes and $\mathcal{A}_v$ denotes the set of edges. Since geographical distance is one of the essential factors to locate neighbor objects, we first create the geometrical relation graph $\mathcal{G}_v^{geo}$ from images by setting edges between an object $v$ and its $M$ nearest objects via $\mathcal{G}_v^{geo} = (\mathcal{V}_v^{geo},\mathcal{A}_v^{geo}) $, $\mathcal{V}_v^{geo} = \{ v_1,...,v_M \}$ and  $\mathcal{A}_v^{geo} = \{ (v,u) \vert u \in \mathcal{V}_v^{geo}, u \neq v \}$. Figure~\ref{fig:graph} (b) shows an example of the geometrical graph.

Besides the geographical distance, we further explore the relationships between objects via scene graph~\cite{johnson2015image}. Scene graph constitutes the semantic relationships between objects. It is a visually-grounded graph over the objects in an image. The nodes denote visual objects and the directed edges depict their pairwise relations. Through the relations excavated via scene graphs, we are able to discover semantic relations between objects in an image that might be physically far apart from each other. It is a necessary complement of the geographical distance. Scene graph uses triplets to represent the relations between two nodes. We denote the relation as a directed edge $a_{s \rightarrow o} = <v_s,r,v_o>$, where $v_s$ denotes the subject object, $v_o$ denotes the target object and $r$ is the relation between them. For example, in the edge $<``man",``playing",``Frisbee">$,  $v_s = ``man"$, $r = ``playing"$ and $v_o = ``Frisbee"$. Figure~\ref{fig:graph} (a) shows an example of the scene graph generated from an image. 

We use scene graph to complement the geographical distance graph with semantic relations. Specifically, we first use a scene graph extractor~\cite{tang2020unbiased} to generate the scene graph $\mathcal{G}_I = (\mathcal{V}_I,\mathcal{A}_I)$ of image $I$, which contains all semantic relationships within the image. Then we generate the semantic graph $\mathcal{G}_v^{sem}$ via $\mathcal{G}_I$ by selecting the edges and nodes from scene graph that are directly or indirectly connected to $v$ as shown in Figure~\ref{fig:graph} (a)-(c). Note that the edges in a semantic graph do not have directions. The resulted semantic graph is denoted as $\mathcal{G}_v^{sem} = (\mathcal{V}_v^{sem}, \mathcal{A}_v^{sem})$ 

Finally, the relation graph is the union of geographical distance graph and scene graph, which is demonstrated in Figure~\ref{fig:graph} (d). 
\begin{align}
\mathcal{G}_v = (\mathcal{V}_v,\mathcal{A}_v) = (\mathcal{V}_v^{geo} \cup \mathcal{V}_v^{sem}, \mathcal{A}_v^{geo} \cup \mathcal{A}_v^{sem})
\end{align}


\subsubsection{Relation Transfer via GNNs}

\begin{figure}
\begin{center}
  \includegraphics[width=0.98\linewidth]{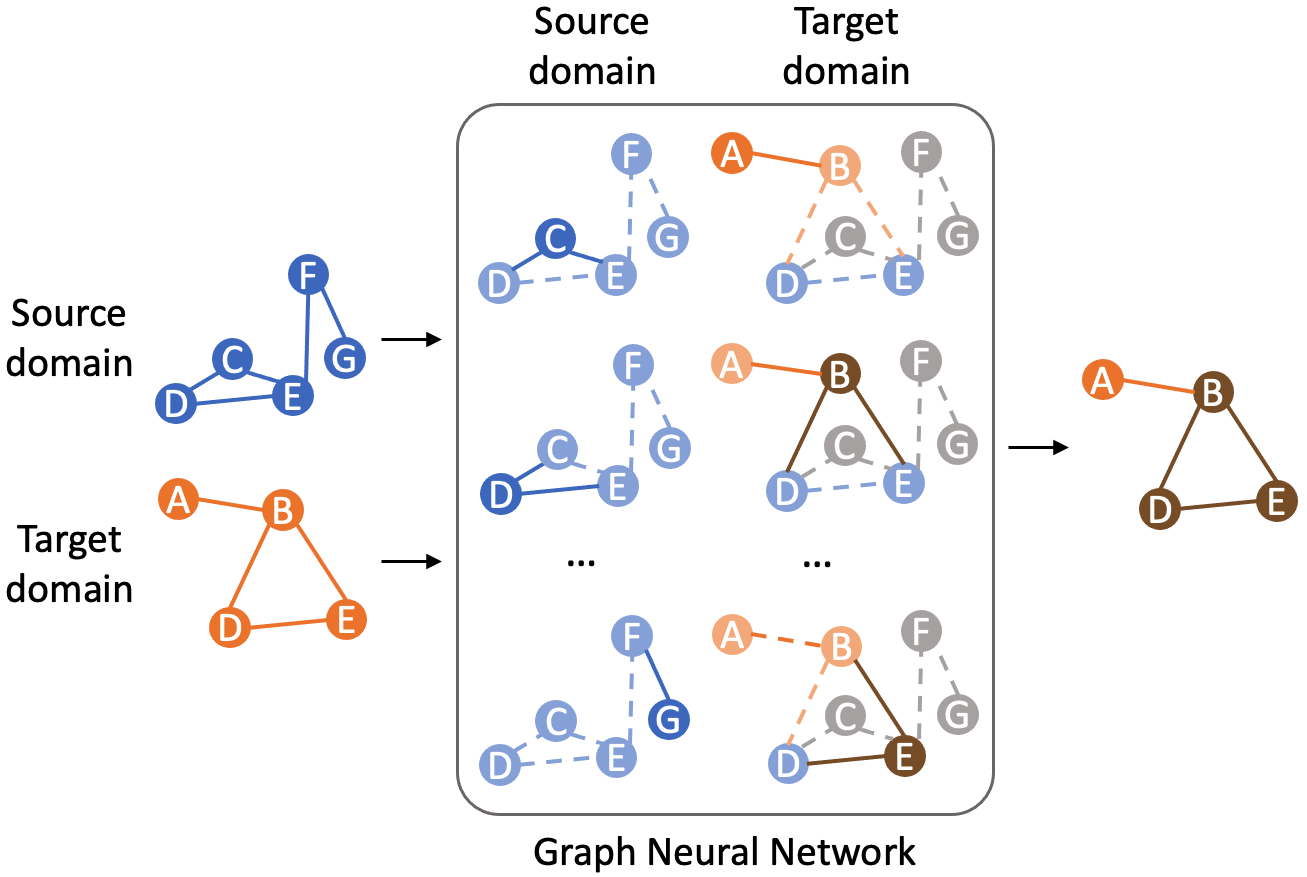}
\end{center}
  \caption{Object relation transfer from source domain to target domain via GNNs. Blue denotes the source domain, orange denotes the target domain and brown denotes that the weights update is affected by both source and target domains. In the GNN, features of each node are updated by aggregating information from connected neighbor nodes. node $D$ and node $E$ are shared by two domains thus the relations between $D$, $E$ and other objects could be transferred to the target domain.}
\label{fig:gcn_graph}
\end{figure}

By aggregating information from neighbor nodes in both source and target domains, Graph Neural Networks (GNN)s can jointly learn an embedding by combing information from the two domains. 
As shown in Figure~\ref{fig:gcn_graph}, the relations between source objects are learned in pre-training, and they are then transferred to the target domain in fine-tuning. 




In the GNN model, each node is naturally defined by its features and the connected nodes. GNN can learn a state embedding $h_v \in \mathbb{R}^{d}$ that contains the information of neighborhood for each node. The state embedding $h_v$ is a $d$-dimension vector of node $v$. $\overline{h_v}$ (the next state of $h_v$) is computed via:
\begin{align}
    \overline{h_v} &= \sigma (\frac{1}{\sum_{u \in N_v}{w_{u \rightarrow v}}}\sum_{u\in N_v}w_{u \rightarrow v} h_u \theta + b ),
\end{align}
where $N_v$ is the neighborhood set of node $v$, $h_u$ is the state embedding of node $u$. $w_{u \rightarrow v}$ is the weighted edge from node $u$ to node $v$. For un-directed graphs, $w_{u \rightarrow v} = w_{v \rightarrow u}$. $\theta$ and $b$ are the weight matrix and bias  of the GNN layer, respectively. $\sigma$ is the activation function. 

We use both the geometrical distance and semantic relation to define the weight matrix $W$. Suppose $D$ denotes the geometrical distance metric, we have:
\begin{align}\begin{aligned}
    &W = [w_{v\rightarrow u}],\ \ (v,u)  \in  \mathcal{A}_v \\
    &w_{v\rightarrow u}  = \left\{\begin{aligned}
    &1, (v,u) \in \mathcal{A}_v^{sem}\\
    &\frac{\min D (\mathcal{A}_v^{geo})}{D(v,u)}, otherwise
    \end{aligned}
    \right.
\end{aligned}\end{align}
where the edges in  $\mathcal{A}_v^{sem}$ are weighted by $1$ and edges in $\mathcal{A}_v^{geo}$ are weighted by their geometrical distance. If an edge exists in both semantic graph and geometrical graph, its weight is calculated based on the semantic graph. 

In the training, we use the C4 feature (last convolutional output of 4th-stage) of Faster R-CNN~\cite{ren2015faster} as the node feature. Besides, we also encode the location of each object in the relation graph with a 5-d vector $l_u = [\frac{x_{tl}}{W},\frac{y_{tl}}{H},\frac{x_{br}}{W},\frac{y_{br}}{H},\frac{w \cdot h}{W\cdot H}]$. $W,H$ are the image weight and height, respectively. Therefore the final feature of node $u$ is the concatenation of the GNN feature $h_u$ and location feature $l_u$. We then apply a fully connected layer to the concatenated feature:
\begin{align}
    r_{u}^{rel} = W_r[h_u;l_u] + b_r
\end{align}
$r_u^{rel}$ is the relation feature of node $u$ in graph $\mathcal{G}_v$.

\subsection{Domain Invariant Modular Learning}

The subject and location modules process the candidate object features, which can be easier transferred between domains. In this paper, we adopt the same structures used in MAttNet~\cite{yu2018mattnet} for the two modules.

\paragraph{Subject module} The subject module processes the visual feature of the candidate object. Specifically, it takes the C3 and C4 features from Faster  R-CNN  as input. C3 feature includes rich space information therefore an attentional pooling~\cite{yu2018mattnet} is applied to the C3 feature thus the salient feature is emphasized. We denote the subject feature of object $v$ as $r_v^{sub}$.

\paragraph{Location module} Similar to the relation module, a 5-d vector is used to encode the location of the candidate object feature. Then a FC layer is used to encode the 5-d vector. The final location feature of object $v$ is $r_v^{loc}$.

\paragraph{Attended language embeddings for visual modules} For each visual module (subject, location and relation), a separate attention is applied to the language expression feature $H$ (see Section \ref{subsec:WordEmbedding}) for an attended embedding: \begin{align}\begin{aligned}
q^m = \sum_{t=1}^T{\frac{\exp(f_m^T h_t) e_t}{\sum_{k=1}^T{\exp(f_m^T h_k)}}}, 
\end{aligned}\end{align}
where $m = \{sub,loc,rel\}$ represents three visual modules. $f_m$ is the learnable weight for each module. $q^m$ are used to attend the feature in visual modules. Besides, the visual modular weights are also learned through a FC layer:
\begin{align}\begin{aligned}
{[w^{sub},w^{loc},w^{rel}]} = \textrm{softmax}(W_m^T[h_0,h_T]+b_m)
\end{aligned}\end{align}

\paragraph{Matching score} For subject and location modules, matching scores $S^{sub}$ and $S^{loc}$ are calculated using cosine similarity (denoted by $M$) between the visual and the expression feature, \ie, $S_v^{sub} = M(r_v^{sub}, q^{sub})$, and $S_v^{loc} = M(r_v^{loc}, q^{loc})$, where $v$ is a candidate object. 
For the relation module, its matching score is the maximum among all nodes in the Relation Graph:
\begin{align}
S_v^{rel} = max_{u \in \mathcal{G}_v, u \neq v}M(r_u^{rel}, q^{rel})
\end{align}
The overall matching score is the weighted average of three visual modules:
\begin{align}
S_v = \sum_{m \in \{sub,loc,rel\}}{w^mS_v^m}
\end{align}
Finally, expression and candidate object pairs (positive and negative) are sampled for a contrastive learning strategy. Hinge loss is applied for optimization. 

\subsection{Revision of Domain Adaptation Approaches}
\label{subsec:existingworks}

Most of the existing domain adaptation algorithms cannot be directly applied to the multi-modal REG problem. We modified several representative domain adaptation methods for the REG problem, which conduct domain adaptation from different perspectives. Notably, some of the recent works such as ensemble-based approaches~\cite{zhou2020domain} are not applicable thus are not included in this paper.
\begin{table*}[t]
\small
\centering
\begin{threeparttable}
\caption{The differences between REG datasets}
\label{tab:dataset}
\begin{tabular}{|c|c|c|c|c|}
\hline
                 & \textbf{RefClef}   & \textbf{RefCOCO}   & \textbf{RefCOCO+}         & \textbf{RefCOCOg} \\ \hline
image from       & ImageCLEF          & MSCOCO             & MSCOCO                   & MSCOCO           \\ \hline
expression style & concise and simple & concise and simple & no location words & complex and long  \\ \hline
expressions/objects/images & 130525/96654/19894 & 142209/50000/19994 & 141564/49856/19992 & 85474/54822/26711  \\ \hline
\end{tabular}
\end{threeparttable}
\end{table*}


\paragraph{Unsupervised Domain Adaptation by Backpropagation (DANN)~\cite{ganin2015unsupervised}} The idea of DANN is to minimize the loss of the label classifier and to maximize the loss of the domain classifier. The latter encourages domain-invariant features to emerge in the course of the optimization. Suppose $\theta_G$ is the parameters in the feature extractor $G$, $\theta_D$ is the parameters in the domain classifier $D$, and $\theta_C$ is the parameters in the matching score module $C$. The overall optimization functional is 
\begin{small}
\begin{align}
   \mathcal{E}(\theta_G,\theta_C,\theta_D) = \sum_{i=1..N,d_i=0}L_y^i(\theta_G,\theta_C)-\lambda\sum_{i=1..N}{L_d^i(\theta_G,\theta_D)},
\end{align}
\end{small}
where $L_y^i$ and$L_d^i$ denote the classification and domain loss, respectively, evaluated at the $i^{th}$ training example. During the training, the parameters $\theta_D$ of the domain classifier and parameters $\theta_G$ and  $\theta_C$ are optimized alternately to minimize the hinge loss and maximize the domain loss. 
In our experiments, 
the DANN algorithm is applied to either/both the visual and language modules. 





\paragraph{Conditional Domain Adversarial Network (CDAN)~\cite{long2017conditional}} Like recent popular domain adaptation algorithms~\cite{you2019universal,hu2020unsupervised}, CDAN requests to involve classification outputs while aligning the features between the source and target domains. In MAttNet, there is an auxiliary task to classify the target object called attribute module $A$. We then use the output from the attribute module to help the domain classifier. The CDAN error terms are:
\begin{small}
\begin{align}
    \mathcal{E}(G,C) = &\sum_{i=1..N,d_i=0}L_y^i(C(G(x_i^i)),y_i^s)\\
    \mathcal{E}(D,G,C) &= - \mathbb{E}_{x_i^s \sim D_s}\log[D(h(G(x_i^s),A(G(x_i^s))))] -\nonumber\\
    &\mathbb{E}_{x_j^t \sim D_t}\log[1-D(h(G(x_j^t),A(G(x_j^t))))]
\end{align}
\end{small}
where $L(.,.)$ is the cross-entropy loss, and $h = G(x) \otimes A(G(x))$ is the joint variable of feature representation $G(x)$ and attribute classifier prediction $A(G(x))$. $\otimes$ denotes the outer product. The method also utilizes a minimax game as the optimization strategy similar to DANN. 


\paragraph{Smooth Representation for unsupervised Domain
Adaptation (SRDA)~\cite{cai2019learning}} Beyond reducing the divergence between source and target domains, it has been studied that the satisfaction of Lipschitz continuity also guarantees an error bound for the target distribution~\cite{ben2014domain}. SRDA proposes the local smooth discrepancy (LSD) term to measure the degree that a sample $x$ breaks down the  local Lipschitz property for Lipschitz continuity~\cite{grandvalet2005semi}. LSD is defined as:
\begin{align}
   LSD(x,\theta) = d(C(G(x)+r), C(G(x))), ||r||\leq \epsilon,
\end{align}
where $d(·, ·)$ is a discrepancy function that measures the divergence between two outputs. $r$ is the noise and $\epsilon$ denotes the maximum norm of $r$. As for the choice of $d(·, ·)$,
we employ MSE loss instead of the cross-entropy loss function used in the original paper since our problem is not a classification problem. 

During the training, all networks are first pre-trained. Then sensitive samples $G(x)+r$ are  generated in the feature space of $G$. Finally, $G$ is optimized to minimize LSD for target samples.
\begin{align}
   \min_{G}&\sum_{i = 1...N}LSD(x_i^t,\theta_G)
\end{align}
\begin{table*}[t]
\caption{Performance comparison (Best results are \textbf{bolded} and the second best are \underline{underlined}.)}
\label{tab:results}
\centering
\begin{threeparttable}
\begin{tabular}{|c|l|ccc|cc|cccc|}
\hline
&Target dataset               & \multicolumn{3}{c|}{RefCOCO+}                                                   & \multicolumn{2}{c|}{RefCOCOg}                   & \multicolumn{4}{c|}{RefClef}                                                                                   \\ 
&{  Split} & {  Val} & {  TestA} & {  TestB} & {  Val} & {  Test} & {  Val} & {  TestA} & {  TestB} & {  TestC} \\ \hline
1&Train on target (full dataset)             & 69.41                      & 73.14                        & 64.12                        & 75.84                      & 75.60                       & 85.51                      & 84.00                        & 88.40                        & 78.08                        \\
2&Finetune on target (full dataset)          & 69.45                      & 72.98                        & 63.88                        & 78.06                      & 78.21                       & 85.69                      & 85.79                        & 89.44                        & 79.35                        \\
3&Train on target              & 57.87                      & 60.25                        & 55.80                        & 67.63                      & 67.44                       & 78.45                      & 75.74                        & 84.12                        & 69.57                        \\
4&Finetune on target           & 59.66                      & 64.41                        & 54.24                        & 69.20                      & 70.56                 &     78.68&77.96&83.60&70.16
                     \\
5&DANN - Visual                     & 60.29                      & 63.08                        & 56.68                        & 69.83                      & 69.94                       & 72.94                      & 73.11                        & 79.54                        & 62.52                        \\
6&DANN - Language                    & 61.31                      & 65.44                        & 56.78                        & 71.81                      & 70.85                       & 78.90                      & 76.60                        & 83.53                        & 72.41                        \\
7&DANN - Language\&Visual                   & 61.77                      & 66.68                        & 57.31                        & 69.24                      & 68.77                       & 73.76                      & 73.53                        & 79.47                        & 66.44                        \\

8&Adversarial-cat           & 61.20                      & 63.60                        & 57.11                        & 69.69                      & 69.08                       & 78.10                      & 77.45                        & 84.34                        & 69.37                        \\
9&CDAN                        & 62.20                      & 66.36                        & 56.88                        & 71.45                      & 70.37                       & 78.29                      & 77.11                        & 83.68                        & 67.32                        \\
10&SRDA                         & 58.35                      & 61.79                        & 53.96                        & 70.83                      & 70.40                       & 72.79                      & 72.68                        & 78.73                        & 65.36                        \\ \hline
11&Ours - Language                    & 64.53                      & 67.90                        & 58.58                        & 73.53                      & 73.15                       & 79.79                      & 79.66                        & 83.83                        &  73.97                        \\
12&Ours - Visual                  & 64.98                      & 68.44                        & 57.66                        & 73.73                      & 73.47                       & 80.05                          & 80.26                        & 84.05                        & \textbf{75.34}                        \\
13&Ours - Relation transfer  & \underline{65.16}                      &\underline{ 68.76}                       & \underline{58.66 }                       & \underline{74.23}                     &\underline{ 73.53 }                    & \underline{80.08 }                    & \underline{80.68}                        & \underline{85.52 }                      & \underline{74.66 }                       \\
14&Ours - Relation transfer + Bert        & \textbf{66.56 }           & \textbf{69.86}           & \textbf{60.50 }          & \textbf{74.49}                     & \textbf{74.10}                   & \textbf{81.12}       & \textbf{82.30 }      & \textbf{86.63}      & 74.36     \\ 

\hline
\end{tabular}
\end{threeparttable}
\end{table*}

\section{Experiments}
\label{sec:experiment}
\subsection{Datasets and Experiment Settings}
The proposed approach is compared with the adapted models presented in Section \ref{subsec:existingworks} on four benchmark datasets: RefClef~\cite{kazemzadeh2014referitgame}, RefCOCO~\cite{yu2016modeling}, RefCOO+~\cite{yu2016modeling} and RefCOCOg~\cite{mao2016generation}. 
The main differences between the four datasets are depicted in Table~\ref{tab:dataset}. We use RefCOCO as the source domain dataset because it has the most number of expressions, and differs from other three datasets from multiple perspectives such as annotating styles or image bases. 


In the experiments, each of the models is first trained with the full RefCOCO data and then finetuned with 10\% of the target data. The evaluation is conducted on the full test data. We implement two baseline models without domain adaption algorithms, one is directly trained on the 10\% target data, and the other is trained on the full RefCOCO data and finetuned on 10\% of the target data. The two baselines act as the lower bounds (rows 3-4 in Table \ref{tab:results}). We also implement two upper-bounds (rows 1-2 in Table \ref{tab:results}), in which we train/finetune on the full target dataset rather than 10\%. For the graph embedding model, we adopt a two-layer GCN in~\cite{kipf2016semi}. We use the pretrained scene graph extractor in~\cite{xu2017scene}. For parameters, we follow the official implementation of these algorithms. An NVIDIA XP GPU is used for all experiments. Best results are recorded among the three different finetuning learning rates, which are equal to, half of, or 10\% of  the pre-training learning rate, respectively. 

\subsection{Results}

Table~\ref{tab:results} shows the results of domain adaptation from RefCOCO to RefCOCO+, RefCOCOg and RefClef. Each of the datsets has different splits. Details about each split can be found in~\cite{kazemzadeh2014referitgame} and ~\cite{yu2016modeling}. The first two rows show the upper-bounds. For DANN~\cite{ganin2015unsupervised}, we consider three scenarios in which the domain classifier is only implemented on the subject domain (-visual), on the language domain (-language) or on both of the visual and language domain (-language\&visual). We also implement one model where we concatenate the features from the visual and language branches and implement a domain classifier proposed in~\cite{tzeng2017adversarial} on top of the concatenation features (adversarial-cat). For this model, a generative adversarial nets (GAN) ~\cite{mirza2014conditional} training strategy is applied. 
Besides, we also include the results of replacing the embedding layer with a pre-trained Bert model~\cite{devlin2018bert} as the language feature extractor. (row 14 in Table~\ref{tab:results}) and using the language and visual relation-based method separately (Table~\ref{tab:results}, rows 11-12).  

Our approach achieves the best results compared with all baselines. Among the three target domains, we have more improvements on RefCOCO+ and RefCOCOg than on RefClef. The main difference is that RefCOCO share the same image base with RefCOCO+ and RefCOCOg but different images with RefClef. It shows that our approch works better while the image base is consistent since our approach focus on the relation-based transfer, which could be potentially combined with image feature transferring method and boost the results further. 


\textbf{Bert embedding.} Bert model~\cite{devlin2018bert} has been widely used in natural language processing tasks. Since Bert model is pre-trained with large language datasets, it improves the quality of language embedding and brings the model more domain-invariant features. Therefore, we investigate to replace the embedding layer in the REG model with the pre-trained Bert model. The results show that it achieves improvements in almost all splits, but not significantly. Besides, for RefClef on split TestC the results with the Bert embedding is slightly worse than without it. That may caused by that TestC contains objects sampled from images that contain at least 2 objects of the same category.~\cite{kazemzadeh2014referitgame}. Bert embedding hardly help distinguish them.  




\textbf{Performance improvement in each modal.} The  three proposed relation-based REG domain adaptation optimization methods could actually be applied separately. We show these results in Table~\ref{tab:results}, rows 11-12. We find that while using these methods alone, they all improve the transferability of the REG model and achieve better results than baselines. But still, the best performance is achieved by stacking them together. Besides, the visual-side relation transfer method (GCN relation encoding) achieves slightly better performance compared with the language-side optimization method (cross-attention word embedding) while being used separately.


\section{Conclusions}
\label{sec:conclusion}

In this paper, we investigate the domain adaptation problem for the REG task, which is still not well studied but widely needed in real-world applications. 
We propose several optimization methods based on the relations on visual and language modalities, which efficiently transfer knowledge learnt in source domain to target domain for the REG problem. Experimental results show that our proposed approach outperforms other domain adaptation methods. We expect our approach from the relation perspective, along with the sound experimental results, can further facilitate future research in this area.

\section{Acknowledgement}
\label{sec:acknow}

This material is based upon work supported by the National Science Foundation under Grant No 1952792.

{
    \small
    \bibliographystyle{ieee_fullname}
    \bibliography{macros,main}
}

\end{document}